\documentclass[runningheads]{llncs}
\usepackage{graphicx}
\usepackage{amsmath,amssymb} 
\usepackage{color}
\usepackage[width=122mm,left=12mm,paperwidth=146mm,height=193mm,top=12mm,paperheight=217mm]{geometry}

\usepackage{epsfig}
\usepackage{algpseudocode}
\usepackage{algorithm}
\usepackage{subfigure}
\usepackage{url}
\usepackage{bbm}
\usepackage[normalem]{ulem}

\newcommand{\argmax}{\operatornamewithlimits{argmax}}
\newcommand{\argmin}{\operatornamewithlimits{argmin}}

\begin{document}

\pagestyle{headings}

\mainmatter
\title{Growing Regression Forests by Classification: Applications to Object Pose Estimation}

\titlerunning{Growing Regression Forests by Classification}

\authorrunning{Kota Hara and Rama Chellappa}

\author{Kota Hara and Rama Chellappa}
\institute{Center for Automation Research, University of Maryland, College Park}

The paper was accepted for publication by ECCV 2014. The title of the paper was changed from ``K-ary Regression Forests for Continuous Pose and Direction Estimation'' to ``Growing Regression Forests by Classification: Applications to Object Pose Estimation.''

\clearpage

\maketitle

\begin{abstract}
In this work, we propose a novel node splitting method for regression trees and incorporate it into the regression forest framework. Unlike traditional binary splitting, where the splitting rule is selected from a predefined set of binary splitting rules via trial-and-error, the proposed node splitting method first finds clusters of the training data which at least locally minimize the empirical loss without considering the input space. Then splitting rules which preserve the found clusters as much as possible are determined by casting the problem into a classification problem. Consequently, our new node splitting method enjoys more freedom in choosing the splitting rules, resulting in more efficient tree structures. In addition to the Euclidean target space, we present a variant which can naturally deal with a circular target space by the proper use of circular statistics. We apply the regression forest employing our node splitting to head pose estimation (Euclidean target space) and car direction estimation (circular target space) and demonstrate that the proposed method significantly outperforms state-of-the-art methods (38.5\% and 22.5\% error reduction respectively).
\end{abstract}

\section{Introduction}

Regression has been successfully applied to various computer vision tasks such as head pose estimation \cite{Haj2012a,Fenzi2013}, object direction estimation \cite{Fenzi2013,Torki2011}, human body pose estimation \cite{Bissacco2007a,Sun2012a,Hara2013} and facial point localization \cite{Dantone2012,Cao2012}, which require continuous outputs. In regression, a mapping from an input space to a target space is learned from the training data. The learned mapping function is used to predict the target values for new data. In computer vision, the input space is typically the high-dimensional image feature space and the target space is a low-dimensional space which represents some high level concepts present in the given image. Due to the complex input-target relationship, non-linear regression methods are usually employed for computer vision tasks.

Among several non-linear regression methods, regression forests \cite{Breiman2001} have been shown to be effective for various computer vision problems \cite{Sun2012a,Criminisi2010,Dantone2012,DFbook}. The regression forest is an ensemble learning method which combines several regression trees \cite{cart} into a strong regressor. The regression trees define recursive partitioning of the input space and each leaf node contains a model for the predictor. In the training stage, the trees are grown in order to reduce the empirical loss over the training data. In the regression forest, each regression tree is independently trained using a random subset of training data and prediction is done by finding the average/mode of outputs from all the trees.


As a node splitting algorithm, binary splitting is commonly employed for regression trees, however, it has limitations regarding how it partitions the input space. The biggest limitation of the standard binary splitting is that a splitting rule at each node is selected by trial-and-error from a predefined set of splitting rules. To maintain the search space manageable, typically simple thresholding operations on a single dimension of the input is chosen. Due to these limitations, the resulting trees are not necessarily efficient in reducing the empirical loss. 

To overcome the above drawbacks of the standard binary splitting scheme, we propose a novel node splitting method and incorporate it into the regression forest framework. In our node splitting method, clusters of the training data which at least locally minimize the empirical loss are first found without being restricted to a predefined set of splitting rules. Then splitting rules which preserve the found clusters as much as possible are determined by casting the problem into a classification problem.   As a by-product, our procedure allows each node in the tree to have more than two child nodes, adding one more level of flexibility to the model. We also propose a way to adaptively determine the number of child nodes at each splitting. Unlike the standard binary splitting method, our splitting procedure enjoys more freedom in choosing the partitioning rules, resulting in more efficient regression tree structures. In addition to the method for the Euclidean target space, we present an extension which can naturally deal with a circular target space by the proper use of circular statistics.

We refer to regression forests (RF) employing our node splitting algorithm as KRF
(K-clusters Regression Forest) and those employing the adaptive determination of the number of child nodes as AKRF. We test KRF and AKRF on Pointing'04 dataset for head pose estimation (Euclidean target space) and EPFL Multi-view Car Dataset for car direction estimation (circular target space) and observe that the proposed methods outperform state-of-the-art with 38.5\% error reduction on Pointing'04 and 22.5\% error reduction on EPFL Multi-view Car Dataset. Also KRF and AKRF significantly outperform other general regression methods including regression forests with the standard binary splitting.

\section{Related work}
A number of inherently regression problems such as head pose estimation and body orientation estimation have been addressed by classification methods by assigning a different pseudo-class label to each of roughly discretized target value (e.g., \cite{Yan2013,Huang2010,Orozco2009,Baltieri2012,Ozuysal2009}). Increasing the number of pseudo-classes allows more precise prediction, however, the classification problem becomes more difficult. This becomes more problematic as the dimensionality of target space increases. In general, discretization is conducted experimentally to balance the desired classification accuracy and precision.

\cite{Weiss1995,Torgo1996} apply k-means clustering to the target space to automatically discretize the target space and assign pseudo-classes. They then solve the classification problem by rule induction algorithms for classification. Though somewhat more sophisticated, these approaches still suffer from problems due to discretization. The difference of our method from approaches discussed above is that in these approaches, pseudo-classes are fixed once determined either by human or clustering algorithms while in our approach, pseudo-classes are \textit{adaptively} redetermined at each node splitting of regression tree training. 

Similarly to our method, \cite{Dobra:2002} converts node splitting tasks into local classification tasks by applying EM algorithm to the joint input-output space. Since clustering is applied to the joint space, their method is not suitable for tasks with high dimensional input space. In fact there experiments are limited to tasks with upto 20 dimensional input space, where their method performs poorly compared to baseline methods.

The work most similar to our method was proposed by Chou \cite{Chou1991} who applied k-means like algorithm to the target space to find a locally optimal set of partitions for regression tree learning. However, this method is limited to the case where the input is a categorical variable. Although we limit ourselves to continuous inputs, our formulation is more general and can be applied to any type of inputs by choosing appropriate classification methods. 



Regression has been widely applied for head pose estimation tasks. \cite{Haj2012a} used kernel partial least squares regression to learn a mapping from HOG features to head poses. Fenzi \cite{Fenzi2013} learned a set of local feature generative model using RBF networks and estimated poses using MAP inference.


A few works considered direction estimation tasks where the direction ranges from 0$^\circ$ and 360$^\circ$. \cite{Herdtweck2013} modified regression forests so that the binary splitting minimizes a cost function specifically designed for direction estimation tasks. \cite{Torki2011} applied supervised manifold learning and used RBF networks to learn a mapping from a point on the learnt manifold to the target space.



\section{Methods}

We denote a set of training data by $\{\mathbf{x}_i,\mathbf{t}_i\}_{i=1}^{N}$ , where $\mathbf{x} \in \mathbb{R}^p$ is an input vector and $\mathbf{t} \in \mathbb{R}^q$ is a target vector. The goal of regression is to learn a function $F^*(\mathbf{x})$ such that the expected value of a certain loss function $\Psi(\mathbf{t},F(\mathbf{x}))$ is minimized:
\begin{equation}\label{eq:30}
F^*(\mathbf{x})=\argmin_{F(\mathbf{x})} \mathrm{E}[\Psi(\mathbf{t},F(\mathbf{x})].
\end{equation}
By approximating the above expected loss by an empirical loss and using the squared loss function, Eq.\ref{eq:30} is reformulated as minimizing the sum of squared errors (SSE):
\begin{equation}\label{eq:50}
F^*(\mathbf{x})=\argmin_{F(\mathbf{x})} \sum_{i=1}^{N} || \mathbf{t}_i - F(\mathbf{x}_i) ||_{2}^{2}.
\end{equation}
However, other loss functions can also be used. In this paper we employ a specialized loss function for a circular target space (Sec.\ref{sec:400}).

In the following subsections, we first explain an abstracted regression tree algorithm, followed by the presentation of a standard binary splitting method normally employed for regression tree training. We then describe the details of our splitting method. An algorithm to adaptively determine the number of child nodes is presented, followed by a modification of our method for the circular target space, which is necessary for direction estimation tasks. Lastly, the regression forest framework for combining regression trees is presented.

\subsection{Abstracted Regression Tree Model}
Regression trees are grown by recursively partitioning the input space into a set of disjoint partitions, starting from a root node which corresponds to the entire input space. At each node splitting stage, a set of splitting rules and prediction models for each partition are determined so as to minimize the certain loss (error). A typical choice for a prediction model is a constant model which is determined as a mean target value of training samples in the partition. However, higher order models such as linear regression can also be used. Throughout this work, we employ the constant model. After each partitioning, corresponding child nodes are created and each training sample is forwarded to one of the child nodes. Each child node is further split if the number of the training samples belonging to that node is larger than a predefined number. 

The essential component of regression tree training is an algorithm for splitting the nodes. Due to the recursive nature of training stage, it suffices to discuss the splitting of the root node where all the training data are available. Subsequent splitting is done with a subset of the training data belonging to each node in exactly the same manner.

Formally, we denote a set of $K$ disjoint partitions of the input space by $\mathcal{R}=\{r_1, r_2, \dots, r_K\}$, a set of constant estimates associated with each partition by $\mathcal{A}=\{\mathbf{a}_1, \dots, \mathbf{a}_K\}$ and the $K$ clusters of the training data by $\mathbf{S}=\{S_1, S_2, \cdots, S_K\}$ where
\begin{equation}\label{eq:55}
S_k = \{i: \mathbf{x}_i \in r_k \}.
\end{equation}

In the squared loss case, a constant estimate, $\mathbf{a}_k$, for the $k$-th partition is computed as the mean target vector of the training samples that fall into $r_k$:
\begin{equation}\label{eq:60}
\mathbf{a}_k = \frac{1}{|S_k|} \sum_{i \in S_k} \mathbf{t}_i.
\end{equation}

The sum of squared errors (SSE) associated with each child node is computed as:
\begin{equation}
\mathrm{SSE}_k=\sum_{i \in S_k} || \mathbf{t}_i - \mathbf{a}_k ||_2^2,
\end{equation}
where $\mathrm{SSE}_k$ is the SSE for the $k$-th child node.
Then the sum of squared errors on the entire training data is computed as:
\begin{equation}\label{eq:150}
\mathrm{SSE}=\sum_{k=1}^{K} \mathrm{SSE}_k = \sum_{k=1}^{K} \sum_{i \in S_k} || \mathbf{t}_i - \mathbf{a}_k ||_2^2.
\end{equation}
The aim of training is to find a set of splitting rules defining the input partitions which minimizes the SSE. 

Assuming there is no further splitting, the regression tree is formally represented as
\begin{equation}\label{eq:100}
H(\mathbf{x};\mathcal{A},\mathcal{R}) = \sum_{k=1}^{K} \mathbf{a}_k \mathbbm{1}(\mathbf{x} \in r_k ),
\end{equation}
where $\mathbbm{1}$ is an indicator function. The regression tree outputs one of the elements of $\mathcal{A}$ depending on to which of the $\mathcal{R}=\{r_1, \dots, r_K\}$, the new data $\mathbf{x}$ belongs. As mentioned earlier, the child nodes are further split as long as the number of the training samples belonging to the node is larger than a predefined number.

\subsection{Standard Binary Node Splitting}
In standard binary regression trees \cite{cart}, $K$ is fixed at two. Each splitting rule is defined as a pair of the index of the input dimension and a threshold. Thus, each binary splitting rule corresponds to a hyperplane that is perpendicular to one of the axes. Among a predefined set of such splitting rules, the one which minimizes the overall SSE (Eq.\ref{eq:150}) is selected by trial-and-error.

The major drawback of the above splitting procedure is that the splitting rules are determined by exhaustively searching the best splitting rule among the predefined set of candidate rules. Essentially, this is the reason why only simple binary splitting rules defined as thresholding on a single dimension are considered in the training stage. Since the candidate rules are severely limited, the selected rules are not necessarily the best among all possible ways to partition the input space. 

\subsection{Proposed Node Splitting}
\begin{figure}[htb]
\begin{center}
\includegraphics[width=3.4in]{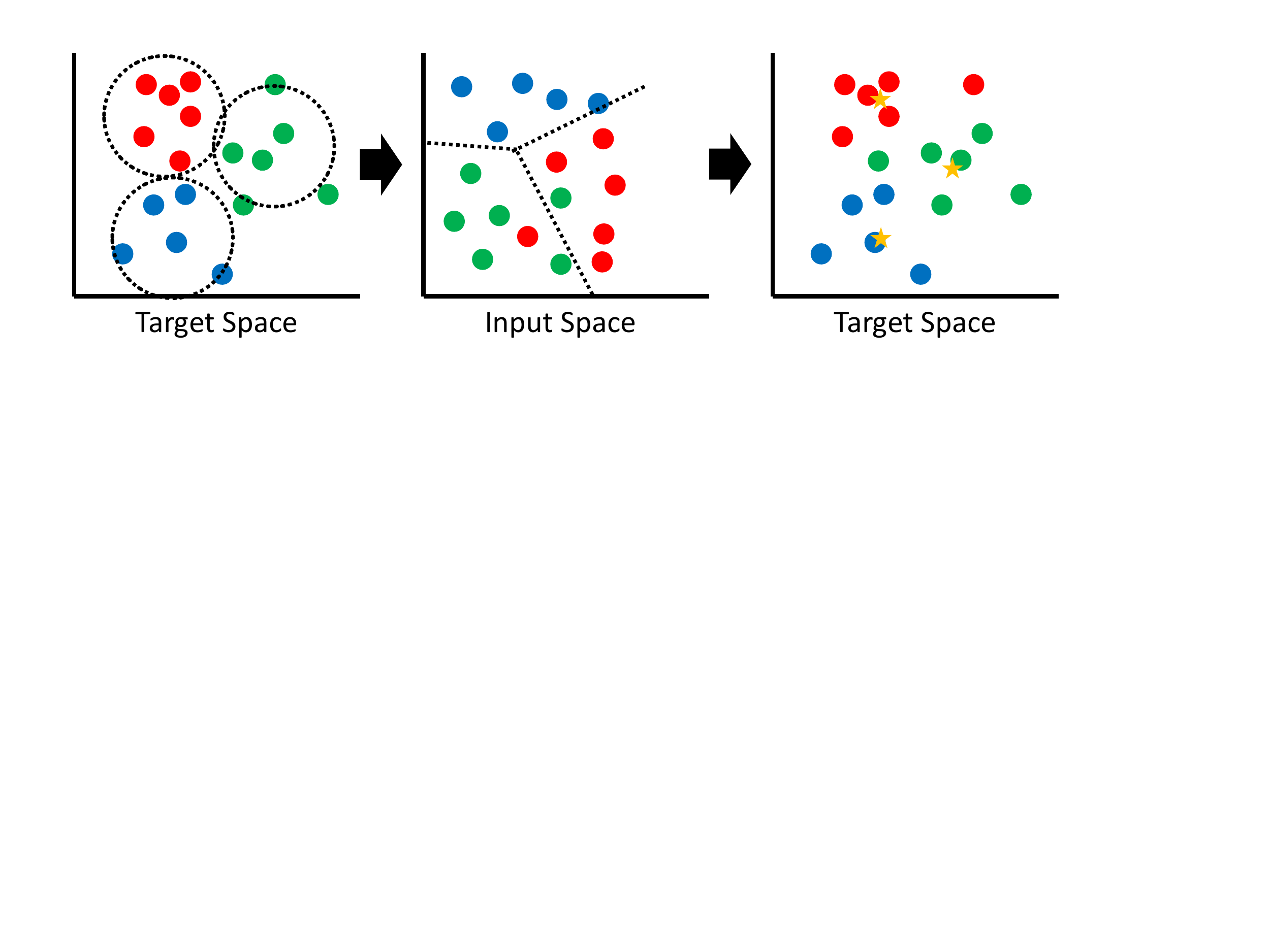}
\caption{An illustration of the proposed splitting method ($K=3$). A set of clusters of the training data are found in the target space by k-means (left). The input partitions preserving the found clusters as much as possible are determined by SVM (middle). If no more splitting is needed, a mean is computed as a constant estimate for each set of colored samples. The yellow stars represent the means. Note that the color of some points change due to misclassification. (right) If further splitting is needed, clusterling is applied to each set of colored samples separately in the target space.  \label{fig:10}}
\end{center}
\end{figure}

In order to overcome the drawbacks of the standard binary splitting procedure, we propose a new splitting procedure which does not rely on trial-and-error. A graphical illustration of the algorithm is given in Fig.\ref{fig:10}. At each node splitting stage, we first find ideal clusters $\mathbf{T}=\{T_1, T_2, \cdots, T_K\}$ of the training data associated with the node, those at least locally minimize the following objective function:
\begin{equation}\label{eq:300}
\min_{\mathbf{T}} \sum_{k=1}^K \sum_{i \in T_k} || \mathbf{t}_i - \mathbf{a}_k ||_{2}^{2}
\end{equation}
where $T_k = \{i : || \mathbf{t}_i - \mathbf{a}_k ||_2 \leq || \mathbf{t}_i - \mathbf{a}_j ||_2, \forall \: 1 \leq j \leq K \}$ and $\mathbf{a}_k = \frac{1}{|T_k|} \sum_{i \in T_k} \mathbf{t}_i$. This minimization can be done by applying the k-means clustering algorithm in the target space with $K$ as the number of clusters. Note the similarity between the objective functions in Eq.\ref{eq:300} and Eq.\ref{eq:150}. The difference is that in Eq.\ref{eq:150}, clusters in $\mathbf{S}$ are indirectly determined by the splitting rules defined in the input space while clusters in $\mathbf{T}$ are directly determined by the k-means algorithm without taking into account the input space. 

After finding $\mathbf{T}$, we find partitions $\mathcal{R}=\{r_1, \dots, r_K\}$ of the input space which preserves $\mathbf{T}$ as much as possible. This task is equivalent to a $K$-class classification problem which aims at determining a cluster ID of each training data based on $\mathbf{x}$. Although any classification method can be used, in this work, we employ L2-regularized L2-loss linear SVM with a one-versus-rest approach. Formally, we solve the following optimization for each cluster using LIBLINEAR \cite{Fan2008}:
 \begin{equation}
\min_{\mathbf{w}_k} ||\mathbf{w}_k||_2 + C \sum_{i=1}^{N} (\max(0,1-l_i^k \mathbf{w}_k^T \mathbf{x}_i ) )^2,
\end{equation}
where $\mathbf{w}_k$ is the weight vector for the $k$-th cluster, $l_i^k = 1$ if $i \in T_k$ and $-1$ otherwise and $C>0$ is a penalty parameter. We set $C=1$ throughout the paper. Each training sample is forwarded to one of the $K$ child nodes by
\begin{equation}\label{eq:310}
k^* = \argmax_{k \in \{1,\cdots,K \}} \mathbf{w}_k^T \mathbf{x}.
\end{equation}

At the last stage of the node splitting procedure, we compute $\mathbf{S}$ (Eq.\ref{eq:55}) and $\mathcal{A}$ (Eq.\ref{eq:60}) based on the constructed splitting rules (Eq.\ref{eq:310}). 

Unlike standard binary splitting, our splitting rules are not limited to hyperplanes that are perpendicular to one of the axes and the clusters are found without being restricted to a set of predefined splitting rules in the input space. Furthermore, our splitting strategy allows each node to have more than two child nodes by employing $K > 2$, adding one more level of flexibility to the model. Note that larger $K$ generally results in smaller value for Eq.\ref{eq:300}, however, since the following classification problem becomes more difficult, the larger $K$ does not necessarily lead to better performance.

\subsection{Adaptive determination of $K$}
Since $K$ is a parameter, we need to determine the value for $K$ by time consuming cross-validation step. In order to avoid the cross-validation step while achieving comparative performance, we propose a method to adaptively determine $K$ at each node based on the sample distribution. 



In this work we employ Bayesian Information Criterion (BIC) \cite{Kashyap1977,Schwarz1978} as a measure to choose $K$. BIC was also used in \cite{Pelleg2000} but with a different formulation. The BIC is designed to balance the model complexity and likelihood. As a result, when a target distribution is complex, a larger number of $K$ is selected and when the target distribution is simple, a smaller value of $K$ is selected. This is in contrast to the non-adaptive method where a fixed number of $K$ is used regardless of the complexity of the distributions.

As k-means clustering itself does not assume any underling probability distribution, we assume that the data are generated from a mixture of isotropic weighted Gaussians with a shared variance. The unbiased estimate for the shared variance is computed as
\begin{equation}\label{eq:320}
\hat{\sigma}^2 = \frac{1}{N-K} \sum_{k=1}^K \sum_{i \in T_k} || \mathbf{t}_i - \mathbf{a}_k ||_{2}^{2}.
\end{equation}

We compute a point probability density for a data point $\mathbf{t}$ belonging to the $k$-th cluster as follows:
\begin{equation}
p(\mathbf{t}) = \frac{|T_k|}{N} \frac{1}{\sqrt{2\pi \hat{\sigma}^2}^q} \exp(- \frac{|| \mathbf{t} - \mathbf{a}_k ||_{2}^{2}}{2\hat{\sigma}^2}).
\end{equation}

Then after simple calculations, the log-likelihood of the data is obtained as
\begin{align}\label{eq:330}
\ln \mathcal{L}(\{\mathbf{t}_i\}_{i=1}^{N}) = \ln \mathbf{\Pi}_{i=1}^{N} p(\mathbf{t}_i) &= \sum_{k=1}^{K} \sum_{i \in T_k} \ln p(\mathbf{t}_i) = \nonumber \\
-\frac{q N}{2} \ln (2 \pi \hat{\sigma}^2 ) - \frac{N-K}{2} & + \sum_{k=1}^{K} |T_k| \ln |T_k| - N \ln N
\end{align}



Finally, the BIC for a particular value of $K$ is computed as
\begin{equation}\label{eq:340}
\mathrm{BIC}_K = -2 \ln \mathcal{L}(\{\mathbf{t}_i\}_{i=1}^{N}) + (K-1+qK+1) \ln N.
\end{equation}

At each node splitting stage, we run the k-means algorithm for each value of $K$ in a manually specified range and select $K$ with the smallest BIC. Throughout this work, we select $K$ from $\{2,3,\dots,40\}$.

\subsection{Modification for a Circular Target Space}\label{sec:400}
1D direction estimation of the object such as cars and pedestrians is unique in that the target variable is periodic, namely, 0$^\circ$ and 360$^\circ$ represent the same direction angle. Thus, the target space can be naturally represented as a unit circle, which is a 1D Riemannian manifold in $\mathit{R}^2$. To deal with a such target space, special treatments are needed since the Euclidean distance is inappropriate. For instance, the distance between 10$^\circ$ and 350$^\circ$ should be shorter than that between 10$^\circ$ and 50$^\circ$ on this manifold.


In our method, such direction estimation problems are naturally addressed by modifying the k-means algorithm and the computation of BIC. The remaining steps are kept unchanged. The k-means clustering method consists of computing cluster centroids and hard assignment of the training samples to the closest centroid. Finding the closest centroid on a circle is trivially done by using the length of the shorter arc as a distance. Due to the periodic nature of the variable, the arithmetic mean is not appropriate for computing the centroids. A typical way to compute the mean of angles is to first convert each angle to a 2D point on a unit circle. The arithmetic mean is then computed on a 2D plane and converted back to the angular value. More specifically, given a set of direction angles $t,\dots, t_N$, the mean direction $a$ is computed by
\begin{equation}\label{eq:660}
a = \mathrm{atan2}( \frac{1}{N} \sum_{i=1}^{N} \sin t_i, \frac{1}{N} \sum_{i=1}^{N} \cos t_i ).
\end{equation}
It is known \cite{DirectionalStatistics2} that $a$ minimizes the sum of a certain distance defined on a circle,
\begin{equation}
a = \argmin_{s} \sum_{i=1}^{N} d(t_i, s)
\end{equation}
where $d(q, s) = 1-\cos(q-s) \in [0,2]$. Thus, the k-means clustering using the above definition of means finds clusters $\mathbf{T}=\{T_1, T_2, \cdots, T_K\}$ of the training data that at least locally minimize the following objective function,
\begin{equation}\label{eq:1300}
\min_{\mathbf{T}} \sum_{k=1}^K \sum_{i \in T_k} ( 1 - \cos( t_i - a_k ) )
\end{equation}
where $T_k = \{i : 1 - \cos(t_i - a_k) \leq 1 - \cos( t_i - a_j ), \forall \: 1 \leq j \leq K \}$.

Using the above k-means algorithm in our node splitting essentially means that we employ distance $d(q,s)$ as a loss function in Eq.\ref{eq:30}. Although squared shorter arc length might be more appropriate for the direction estimation task, there is no constant time algorithm to find a mean which minimizes it. Also as will be explained shortly, the above definition of the mean coincides with the maximum likelihood estimate of the mean of a certain probability distribution defined on a circle.

As in the Euclidean target case, we can also adaptively determine the value for $K$ at each node using BIC. As a density function, the Gaussian distribution is not appropriate. A suitable choice is the von Mises distribution, which is a periodic continuous probability distribution defined on a circle,
\begin{equation}
p(t| a, \kappa ) = \frac{1}{2 \pi I_0(\kappa)} \exp{( \kappa \cdot \cos( t - a ))}
\end{equation}
where $a$, $\kappa$ are analogous to the mean and variance of the Gaussian distribution and $I_{\lambda}$ is the modified Bessel function of order $\lambda$. It is known \cite{CircularStatistics} that the maximum likelihood estimate of $a$ is computed by Eq.\ref{eq:660} and that of $\kappa$ satisfies
\begin{equation}
\frac{I_{1}(\kappa)}{I_{0}(\kappa)} = \sqrt{ ( \frac{1}{N} \sum_{i=1}^{N} \sin t_i )^2 + (\frac{1}{N} \sum_{i=1}^{N} \cos t_i )^2 } = \frac{1}{N} \sum_{i=1}^{N} \cos( t_i - a).
\end{equation}
Note that, from the second term, the above quantity is the Euclidean norm of the mean vector obtained by converting each angle to a 2D point on a unit circle.

Similar to the derivation for the Euclidean case, we assume that the data are generated from a mixture of weighted von Mises distributions with a shared $\kappa$. The mean $a_k$ of k-th von Mises distribution is same as the mean of the k-th cluster obtained by the k-means clustering. The shared value for $\kappa$ is obtained by solving the following equation
\begin{equation}
\frac{I_{1}(\kappa)}{I_{0}(\kappa)} = \frac{1}{N} \sum_{k=1}^{K} \sum_{i \in T_k} \cos( t_i - a_k).
\end{equation}

Since there is no closed form solution for the above equation, we use the following approximation proposed in \cite{DirectionalStatistics},
\begin{equation}
\kappa \approx \frac{1}{2(1-\frac{I_{1}(\kappa)}{I_{0}(\kappa)})}.
\end{equation}

Then, a point probability density for a data point $t$ belonging to the k-th cluster is computed as:
\begin{equation}
p(t | a_k, \kappa ) = \frac{|T_k|}{N} \frac{\exp{( \kappa \cdot \cos( t - a_k ))}}{2 \pi I_0(\kappa)}.
\end{equation}
After simple calculations, the log-likelihood of the data is obtained as
\begin{align}
\ln \mathcal{L}(\{t_i\}_{i=1}^{N}) &= \ln \Pi_{i=1}^{N} p(t_i) = \sum_{k=1}^{K} \sum_{i \in T_k} \ln p(t_i) = \nonumber \\
- N \ln(2\pi I_0(\kappa)) &+ \kappa \sum_{k=1}^{K} \sum_{i \in T_k} \cos( t_i - a_k ) + \sum_{k=1}^{K} |T_k| \ln |T_k| - N \ln N.
\end{align}

Finally, the BIC for a particular value of $K$ is computed as
\begin{equation}\label{eq:740}
\mathrm{BIC}_K = -2 \ln \mathcal{L}(\{t_i\}_{i=1}^{N}) + 2K \ln N.
\end{equation}
where the last term is obtained by putting $q=1$ into the last term of Eq.\ref{eq:340}. 

\subsection{Regression Forest} \label{sec:500}
We use the regression forest \cite{Breiman2001} as the final regression model. The regression forest is an ensemble learning method for regression which first constructs multiple regression trees from random subsets of training data. Testing is done by computing the mean of the outputs from each regression tree. We denote the ratio of random samples as $\beta \in (0, 1.0]$. For the Euclidean target case, arithmetic mean is used to obtain the final estimate and for the circular target case, the mean defined in Eq.\ref{eq:660} is used.


For the regression forest with standard binary regression trees, an additional randomness is typically injected. In finding the best splitting function at each node, only a randomly selected subset of the feature dimensions is considered. We denote the ratio of randomly chosen feature dimensions as $\gamma \in (0, 1.0]$. For the regression forest with our regression trees, we always consider all feature dimensions. However, another form of randomness is naturally injected by randomly selecting the data points as the initial cluster centroids in the k-means algorithm.

\section{Experiments}\label{sec:experiments}

\subsection{Head Pose Estimation}
We test the effectiveness of KRF and AKRF for the Euclidean target space on the head pose estimation task. We adopt Pointing'04 dataset \cite{Gourier2004}. The dataset contains head images of 15 subjects and for each subject there are two series of 93 images with different poses represented by pitch and yaw.



The dataset comes with manually specified bounding boxes indicating the head regions. Based on the bounding boxes, we crop and resize the image patches to $64 \times 64$ pixels image patches and compute multiscale HOG from each image patch with cell size 8, 16, 32 and $2\times2$ cell blocks. The orientation histogram for each cell is computed with signed gradients for 9 orientation bins. The resulting HOG feature is 2124 dimensional. 


First, we compare the KRF and AKRF with other general regression methods using the same image features. We choose standard binary regression forest (BRF) \cite{Breiman2001}, kernel PLS \cite{Rosipal2001} and $\epsilon$-SVR with RBF kernels \cite{SVRbook}, all of which have been widely used for various computer vision tasks. The first series of images from all subjects are used as training set and the second series of images are used for testing. The performance is measured by Mean Absolute Error in degree. For KRF, AKRF and BRF, we terminate node splitting once the number of training data associated with each leaf node is less than 5. The number of trees combined is set to 20. $K$ for KRF, $\beta$ for KRF, AKRF and BRF and $\gamma$ for BRF are all determined by 5-fold cross-validation on the training set. For kernel PLS, we use the implementation provided by the author of \cite{Rosipal2001} and for $\epsilon$-SVR, we use LIBSVM package \cite{Chang2011a}. All the parameters for kernel PLS and $\epsilon$-SVR are also determined by 5-fold cross-validation. As can been seen in Table \ref{tb:1100}, both KRF and AKRF work significantly better than other regression methods. Also our methods are computationally efficient (Table \ref{tb:1100}). KRF and AKRF take only 7.7 msec and 8.7 msec, respectively, to process one image including feature computation with a single thread. 
 
 
\begin{table}[htb]
\begin{center}
  \caption{MAE in degree of different regression methods on the Pointing'04 dataset (even train/test split). Time to process one image including HOG computation is also shown.\label{tb:1100}}
  \begin{tabular}{|c||c|c|c||c|} \hline
   Methods & yaw & pitch & average & testing time (msec) \\ \hline
   \textbf{KRF} & 5.32 & 3.52 & 4.42 & 7.7 \\ \hline
   \textbf{AKRF} & 5.49 & 4.18 & 4.83 & 8.7 \\ \hline
   BRF \cite{Breiman2001} & 7.77 & 8.01 & 7.89 & 4.5 \\ \hline
   Kernel PLS \cite{Rosipal2001} & 7.35 & 7.02 & 7.18 & 86.2 \\ \hline 
   $\epsilon$-SVR \cite{SVRbook} & 7.34 & 7.02 & 7.18 & 189.2 \\ \hline 
   \end{tabular}
\end{center}
\end{table}

Table \ref{tb:1200} compares KRF and AKRF with prior art. Since the previous works report the 5-fold cross-validation estimate on the whole dataset, we also follow the same protocol. KRF and AKRF advance state-of-the-art with 38.5\% and 29.7\%  reduction in the average MAE, respectively.

\begin{table}[htb]
\begin{center}
  \caption{Head pose estimation results on the Pointing'04 dataset (5-fold cross-validation)\label{tb:1200}}
  \begin{tabular}{|c||c|c|c|} \hline
   & yaw & pitch & average \\ \hline
   \textbf{KRF} & 5.29 & 2.51 & 3.90 \\ \hline
   \textbf{AKRF} & 5.50 & 3.41 & 4.46 \\ \hline
   Fenzi \cite{Fenzi2013} & 5.94 & 6.73 & 6.34  \\ \hline
   Haj \cite{Haj2012a} Kernel PLS & 6.56 & 6.61 & 6.59 \\ \hline
   Haj \cite{Haj2012a} PLS & 11.29 & 10.52 & 10.91 \\ \hline 
   \end{tabular}
\end{center}
\end{table}

\begin{figure}[htb]
\begin{center}
\includegraphics[width=3.0in]{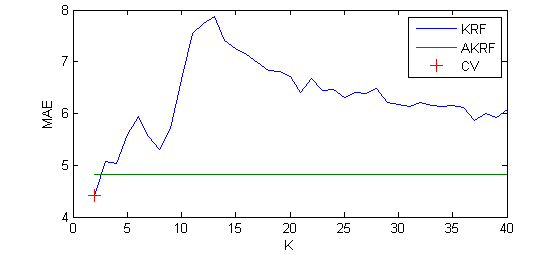}
\caption{Pointing'04: The effect of $K$ of KRF on the average MAE. ``CV'' indicates the value of KRF selected by cross-validation. \label{fig:100}}
\end{center}
\end{figure}

\begin{figure}[htb]
\begin{center}
\includegraphics[width=3.2in]{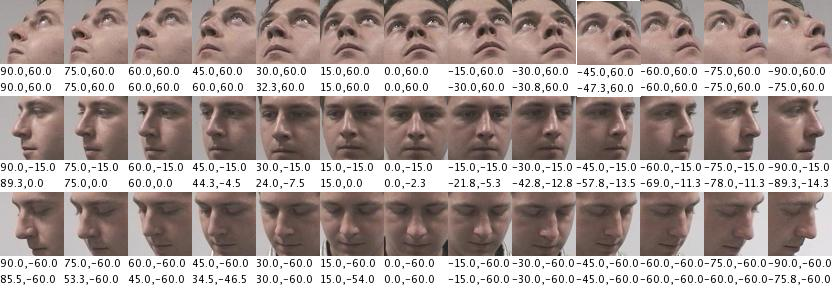}
\caption{Some estimation results of the second sequence of person 13. The top numbers
are the ground truth yaw and pitch and the bottom numbers are the estimated yaw and pitch.\label{fig:110}}
\end{center}
\end{figure}

Fig.\ref{fig:100} shows the effect of $K$ of KRF on the average MAE along with the average MAE of AKRF. In this experiment, the cross-validation process successfully selects $K$ with the best performance. AKRF works better than KRF with the second best $K$. The overall training time is much faster with AKRF since the cross-validation step for determining the value of $K$ is not necessary. To train a single regression tree with $\beta=1$, AKRF takes only 6.8 sec while KRF takes 331.4 sec for the cross-validation and 4.4 sec for training a final model. As a reference, BRF takes 1.7 sec to train a single tree with $\beta=1$ and $\gamma=0.4$. Finally, some estimation results by AKRF on the second sequence of person 13 are shown in Fig.\ref{fig:110}.

\subsection{Car Direction Estimation}
We test KRF and AKRF for circular target space (denoted as KRF-circle and AKRF-circle respectively) on the EPFL Multi-view Car Dataset \cite{Ozuysal2009}. The dataset contains 20 sequences of images of cars with various directions. Each sequence contains images of only one instance of car. In total, there are 2299 images in the dataset. Each image comes with a bounding box specifying the location of the car and ground truth for the direction of the car. The direction ranges from 0$^\circ$ to 360$^\circ$.  As input features, multiscale HOG features with the same parameters as in the previous experiment are extracted from $64 \times 64$ pixels image patches obtained by resizing the given bounding boxes.



The algorithm is evaluated by using the first 10 sequences for training and the remaining 10 sequences for testing. In Table \ref{tb:103}, we compare the KRF-circle and AKRF-circle with previous work. We also include the performance of BRF, Kernel PLS and $\epsilon$-SVR with RBF kernels using the same HOG features. For BRF, we extend it to directly minimize the same loss function ($d(q,s)=1-\cos(q-s)$) as with KRF-circle and AKRF-circle (denoted by BRF-circle). For Kernel PLS and $\epsilon$-SVR, we first map direction angles to 2d points on a unit circle and train regressors using the mapped points as target values. In testing phase, a 2d point coordinate $(x,y)$ is first estimated and then mapped back to the angle by $\mathrm{atan2}(y,x)$. All the parameters are determined by leave-one-sequence-out cross-validation on the training set. The performance is evaluated by the Mean Absolute Error (MAE) measured in degrees. In addition, the MAE of 90-th percentile of the absolute errors and that of 95-th percentile are reported, following the convention from the prior works.

\begin{figure}[htb]
\begin{center}
\includegraphics[width=3.2in]{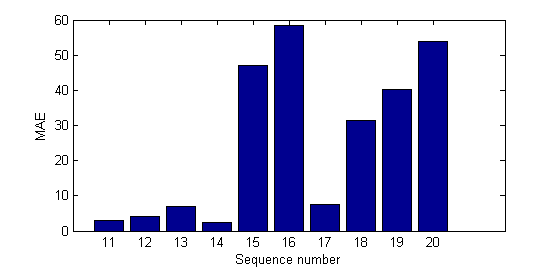}
\caption{MAE of AKRF computed on each sequence in the testing set \label{fig:210}}
\end{center}
\end{figure}

\begin{figure}[htb]
\begin{center}
\includegraphics[width=3.2in]{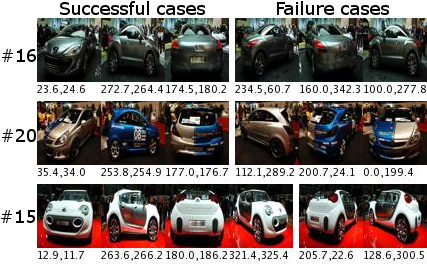}
\caption{Representative results from the \textit{worst} three sequences in the testing set. The numbers under each image are the ground truth direction (left) and the estimated direction (right). Most of the failure cases are due to the flipping error. \label{fig:220}}
\end{center}
\end{figure}

\begin{figure}[htb]
\begin{center}
\includegraphics[width=3.0in]{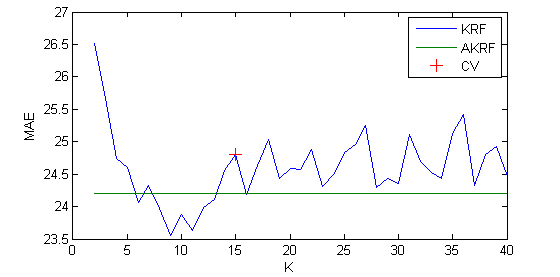}
\caption{EPFL Multi-view Car: The effect of $K$ of KRF on MAE. ``CV'' indicates the value of KRF selected by cross-validation.  \label{fig:250}}
\end{center}
\end{figure}

As can be seen from Table \ref{tb:103}, both KRF-circle and AKRF-circle work much better than existing regression methods. In particular, the improvement over BRF-circle is notable. Our methods also advance state-of-the-art with 22.5\% and 20.7\% reduction in MAE from the previous best method, respectively. In Fig.\ref{fig:210}, we show the MAE of AKRF-circle computed on each sequence in the testing set. The performance varies significantly among different sequences (car models). Fig.\ref{fig:220} shows some representative results from the \textit{worst} three sequences in the testing set (seq 16, 20 and 15). We notice that most of the failure cases are due to the flipping errors ($\approx 180^\circ$) which mostly occur at particular intervals of directions. Fig.\ref{fig:250} shows the effect of $K$ of KRF-circle. The performance of the AKRF-circle is comparable to that of KRF-circle with $K$ selected by the cross-validation.

\begin{table}[htb]
\begin{center}
  \caption{Car direction estimation results on the EPFL Multi-view Car Dataset\label{tb:103}}
  \begin{tabular}{|c|c|c|c|	} \hline
   Method & MAE ($^\circ$) 90-th percentile & MAE ($^\circ$) 95-th percentile & MAE ($^\circ$) \\ \hline
   \textbf{KRF-circle} & 8.32 & 16.76 & 24.80 \\ \hline
   \textbf{AKRF-circle} & 7.73 & 16.18 & 24.24 \\ \hline
   BRF-circle & 23.97 & 30.95 & 38.13 \\ \hline
   Kernel PLS & 16.86 & 21.20 & 27.65 \\ \hline
   $\epsilon$-SVR & 17.38 & 22.70 & 29.41 \\ \hline \hline
   Fenzi et al. \cite{Fenzi2013} & 14.51 & 22.83 & 31.27 \\ \hline
   Torki et al. \cite{Torki2011} & 19.4 & 26.7 & 33.98 \\ \hline
   Ozuysal et al. \cite{Ozuysal2009} & - & - & 46.48 \\ \hline   
   \end{tabular}
\end{center}
\end{table}

\section{Conclusion}
In this paper, we proposed a novel node splitting algorithm for regression tree training. Unlike previous works, our method does not rely on a trial-and-error process to find the best splitting rules from a predefined set of rules, providing more flexibility to the model. Combined with the regression forest framework, our methods work significantly better than state-of-the-art methods on head pose estimation and car direction estimation tasks.

\noindent\textbf{Acknowledgements.} This research was supported by a MURI grant from the US
Office of Naval Research under N00014-10-1-0934.

\clearpage

\bibliographystyle{splncs03}
\bibliography{library,others}
\end{document}